\documentclass[pdflatex,sn-mathphys-num]{sn-jnl}

\usepackage{amsmath}
\usepackage{graphicx}
\usepackage{booktabs}
\usepackage{xcolor}
\usepackage{float}

\graphicspath{{figures/}}

\begin{document}

\title[TutlAit v1 Tamazight speech dataset]{TutlAit v1: a crowdsourced Moroccan Tamazight speech dataset with Arabic transcriptions and regional accent labels}

\author*[1]{\fnm{Mohamed-Amine} \sur{Chadi}}\email{m.chadi@uca.ac.ma}

\author[1]{\fnm{Ezzahra} \sur{Ait El Arbi}}

\author[1]{\fnm{Ismail} \sur{Khayoub}}

\author[1]{\fnm{Aymane} \sur{Fadili}}

\author[1]{\fnm{Yassine} \sur{Ennhili}}

\author[1]{\fnm{Jadjigua} \sur{Bouali}}

\author[1]{\fnm{Hanane} \sur{Inhid}}

\author[1]{\fnm{Mohammed} \sur{Ameksa}}

\author[1]{\fnm{Hajar} \sur{Mousannif}}

\affil*[1]{\orgdiv{Department of Computer Science, Faculty of Sciences Semlalia}, \orgname{Cadi Ayyad University}, \orgaddress{\street{Boulevard Prince My Abdellah, B.P.\ 2390}, \city{Marrakech}, \postcode{40000}, \country{Morocco}}}

\abstract{Tamazight (Amazigh) is, together with Arabic, one of the two official languages of Morocco, yet it remains severely under-resourced for speech technology: publicly available labelled audio is scarce, generally lacks information on the regional variety spoken, and is often of uneven transcription quality. This article describes the TutlAit dataset, a corpus of Moroccan Tamazight speech paired with Modern Standard Arabic text and explicit regional accent labels. The data were collected with TutlAit, a purpose-built crowdsourcing web application (React 18 front-end, Django 5 / Django REST Framework back-end, PostgreSQL database). Native speakers recruited through targeted LinkedIn and Instagram campaigns created an account, declared their regional variety (Atlas, Souss, Rif or other) and demographic information, and then contributed through two workflows: Text-to-Audio, in which an Arabic sentence is displayed and the volunteer records its oral Tamazight rendering in the browser, and Audio-to-Text, in which a Tamazight excerpt is played and the volunteer types its Arabic transcription. A complementary set of segments was obtained from freely accessible Tamazight audiovisual media, segmented and annotated with ELAN and imported through a bulk CSV/ZIP pipeline. Every upload is converted server-side to 16\,kHz mono WAV, hashed with SHA-256 for duplicate rejection, checked for duration bounds and validated by an administrator. The dataset contains 13{,}384 audio files totalling 75{,}231 seconds (approximately 20.9 hours, about 3.01\,GB). The Atlas variety accounts for 9{,}956 files (14.08\,h) and the Souss variety for 3{,}378 files (6.75\,h); small Rif (22 files) and Kabyle (28 files) subsets are also included. The corpus can be reused for speech recognition, speech translation and accent identification for Moroccan Tamazight.}


\maketitle

\noindent\textbf{Note.}
The data volume reported in this article is about 20 hours of validated speech. A further 5 hours are still being collected and checked. Each time new validated recordings are added, they are published in the same Hugging Face dataset: \url{https://huggingface.co/datasets/Ma-OpenHub/Tutlait-v1}.

\section{Introduction}\label{sec:intro}

Morocco is officially bilingual, with Arabic and Tamazight recognised in the 2011 Constitution~\cite{morocco2011constitution}. According to the 2024 national census, Tamazight is spoken by about 24.8\% of the population, distributed across three main regional varieties: Tachelhit in the Souss (14.2\%), Tamazight of the Middle Atlas (7.4\%) and Tarifit in the Rif (3.2\%)~\cite{hcp2024amazigh}. Despite this, speech technology for Tamazight lags far behind Arabic: state-of-the-art multilingual systems were not trained on the language~\cite{radford2023whisper}, community initiatives have collected only a few hours of speech~\cite{oktem2025awal}, and the most accessible public dataset~\cite{dhimi2024tamazight} has no accent labels and contains transcription and audio-quality issues.

The TutlAit dataset was compiled to build a Tamazight speech-to-Arabic-text corpus large enough to fine-tune modern speech recognition models, while explicitly recording the regional variety of every speaker. To reach speakers across Morocco, a dedicated crowdsourcing web application, TutlAit, was designed and deployed. This article documents the resulting corpus, its structure and the collection methodology so that it can be reused and extended.

The dataset is of value to the research community for the following reasons.
\begin{itemize}
\item Tamazight is an official language of Morocco spoken by roughly a quarter of the population~\cite{hcp2024amazigh}, yet it is almost absent from large multilingual speech models such as Whisper~\cite{radford2023whisper}. This dataset provides about 20.9 hours of labelled Moroccan Tamazight speech, a volume that is rarely available publicly for this language.
\item Unlike existing public Tamazight resources~\cite{dhimi2024tamazight}, every recording is tagged with the regional variety declared by the speaker (Atlas, Souss, Rif, Kabyle). This makes the corpus suitable for per-accent evaluation, dialect identification and studies of acoustic and lexical variation across Amazigh varieties.
\item Each audio clip is paired with Modern Standard Arabic text. The corpus can therefore be used both for Tamazight speech recognition with Arabic-script output and for end-to-end Tamazight-to-Arabic speech translation, two tasks for which parallel speech--text data are extremely scarce.
\item All files share a uniform format (16\,kHz mono WAV) and are indexed in a single tabular file with duration, accent and anonymised speaker attributes, so the data can be loaded directly with standard speech toolkits without further preprocessing.
\item The collection protocol, quality-control rules and the TutlAit platform described here can be reused to extend the corpus to under-represented varieties (notably Rif) or to other low-resource languages~\cite{akallouch2025survey,oktem2025awal}.
\end{itemize}

\section{Data description}\label{sec:data}

The dataset~\cite{aitelarbi2026tutlait} is distributed as one compressed audio archive and one metadata index:
\begin{itemize}
\item \texttt{tutlait\_audio.zip} contains all 13{,}384 audio files, grouped by regional variety in the sub-folders \texttt{audio/atlas/}, \texttt{audio/souss/}, \texttt{audio/rif/} and \texttt{audio/kabyle/}. Each file is a 16\,kHz, mono, 16-bit PCM WAV named \texttt{<annotation\_id>.wav}.
\item \texttt{metadata.csv} (and an identical \texttt{metadata.xlsx}) is a UTF-8 index with one row per audio file. Its columns are listed in Table~\ref{tab:columns}.
\item \texttt{stats/} holds the descriptive figures reproduced in this article.
\item \texttt{README.md} gives the licence, the citation and a short loading example.
\end{itemize}

\begin{table}[ht]
\caption{Columns of the metadata index \texttt{metadata.csv}.}\label{tab:columns}
\begin{tabular}{@{}p{2.6cm}p{1.8cm}p{7.2cm}@{}}
\toprule
Column & Type & Description \\
\midrule
\texttt{annotation\_id} & integer & Unique identifier of the contribution; matches the WAV file name. \\
\texttt{audio\_path} & string & Relative path of the WAV file inside \texttt{tutlait\_audio.zip}. \\
\texttt{text} & string & Modern Standard Arabic text (source sentence for Text-to-Audio, transcription for Audio-to-Text). \\
\texttt{accent} & categorical & Regional variety: \texttt{atlas}, \texttt{souss}, \texttt{rif}, \texttt{kabyle}. \\
\texttt{mode} & categorical & Workflow: \texttt{text\_to\_audio}, \texttt{audio\_to\_text} or \texttt{bulk\_import}. \\
\texttt{duration\_s} & float & Clip duration in seconds. \\
\texttt{sample\_rate} & integer & Sampling rate (always 16000). \\
\texttt{speaker\_id} & string & Anonymised speaker identifier (hash of the account identifier). \\
\texttt{gender} & categorical & Self-declared gender (\texttt{female}, \texttt{male}, \texttt{unspecified}). \\
\texttt{age\_group} & categorical & Self-declared age bracket. \\
\texttt{region} & string & Self-declared region of origin. \\
\texttt{sha256} & string & SHA-256 hash of the audio file. \\
\texttt{status} & categorical & Review outcome (\texttt{validated}). Only validated contributions are released. \\
\botrule
\end{tabular}
\end{table}

Table~\ref{tab:global_stats} summarises the corpus: 13{,}384 audio files, a cumulative duration of 75{,}231 seconds (approximately 20.9 hours) and a total size of about 3.01\,GB.

\begin{table}[ht]
\caption{Global statistics of the TutlAit dataset.}\label{tab:global_stats}
\begin{tabular}{@{}lr@{}}
\toprule
Metric & Value \\
\midrule
Number of audio files & 13{,}384 \\
Total duration (seconds) & 75{,}231 \\
Total duration (hours) & $\approx$ 20.9 h \\
Total size & $\approx$ 3.01 GB \\
Mean duration per file & 5.62 s \\
Median duration per file & 4.03 s \\
Standard deviation of durations & 5.32 s \\
Minimum duration & 0.19 s \\
Maximum duration & 120.26 s \\
Audio format & WAV, 16 kHz, mono, 16-bit PCM \\
Text language / script & Modern Standard Arabic \\
\botrule
\end{tabular}
\end{table}

\begin{figure}[ht]
\centering
\includegraphics[width=0.85\linewidth]{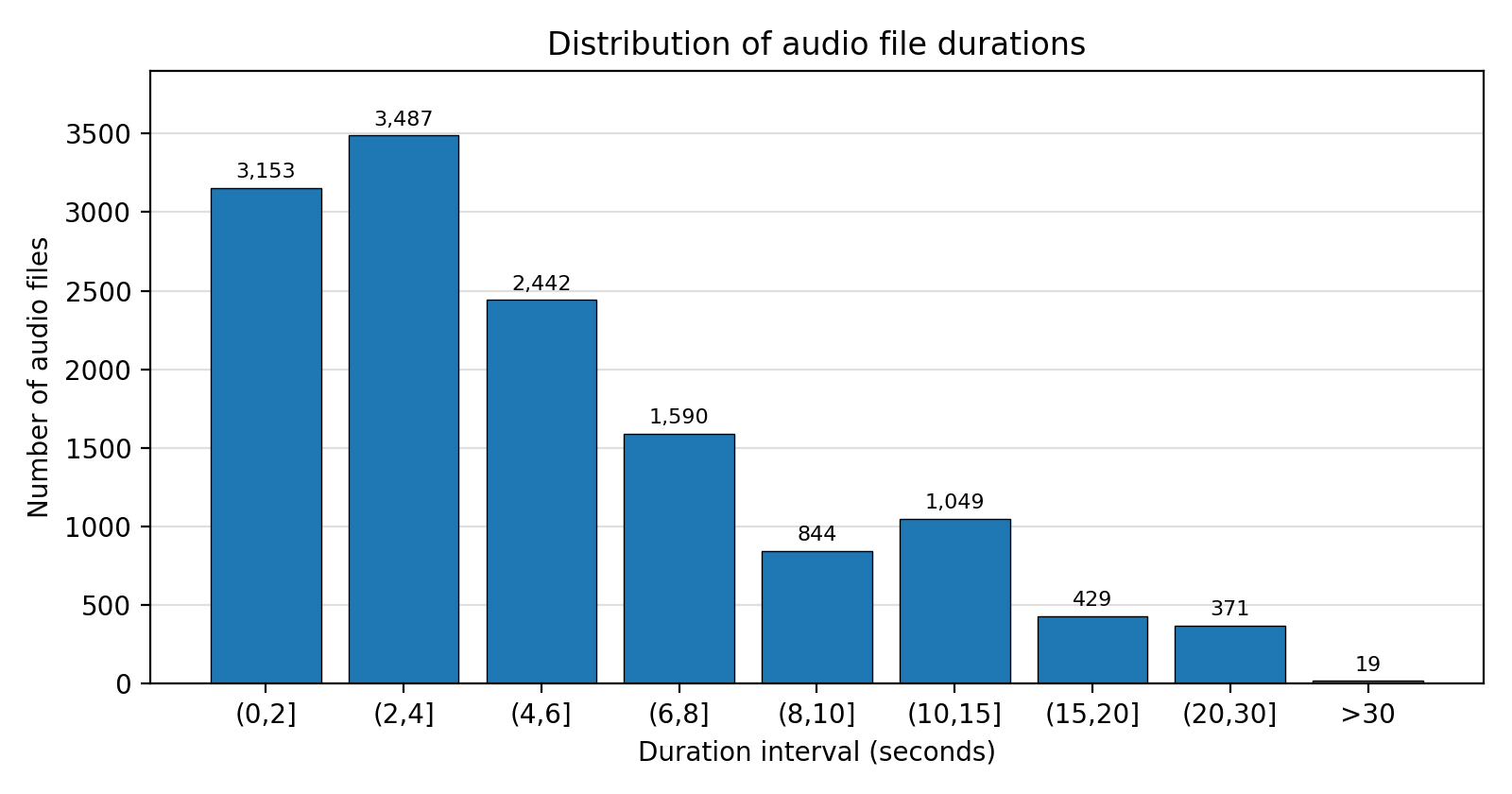}
\caption{Distribution of audio file durations in the TutlAit dataset.}\label{fig:duration_dist}
\end{figure}

Table~\ref{tab:duration_dist} and Fig.~\ref{fig:duration_dist} give the distribution of clip durations. About half of the recordings are shorter than 4 seconds, which corresponds to sentence-level recordings; 19 files (0.14\%) exceed 30 seconds.

\begin{table}[ht]
\caption{Distribution of audio files by duration.}\label{tab:duration_dist}
\begin{tabular}{@{}crr@{}}
\toprule
Duration interval (s) & Number of files & Percentage \\
\midrule
(0, 2] & 3{,}153 & 23.56\% \\
(2, 4] & 3{,}487 & 26.05\% \\
(4, 6] & 2{,}442 & 18.25\% \\
(6, 8] & 1{,}590 & 11.88\% \\
(8, 10] & 844 & 6.31\% \\
(10, 15] & 1{,}049 & 7.84\% \\
(15, 20] & 429 & 3.21\% \\
(20, 30] & 371 & 2.77\% \\
$> 30$ & 19 & 0.14\% \\
\midrule
Total & 13{,}384 & 100\% \\
\botrule
\end{tabular}
\end{table}

\begin{figure}[ht]
\centering
\includegraphics[width=0.72\linewidth]{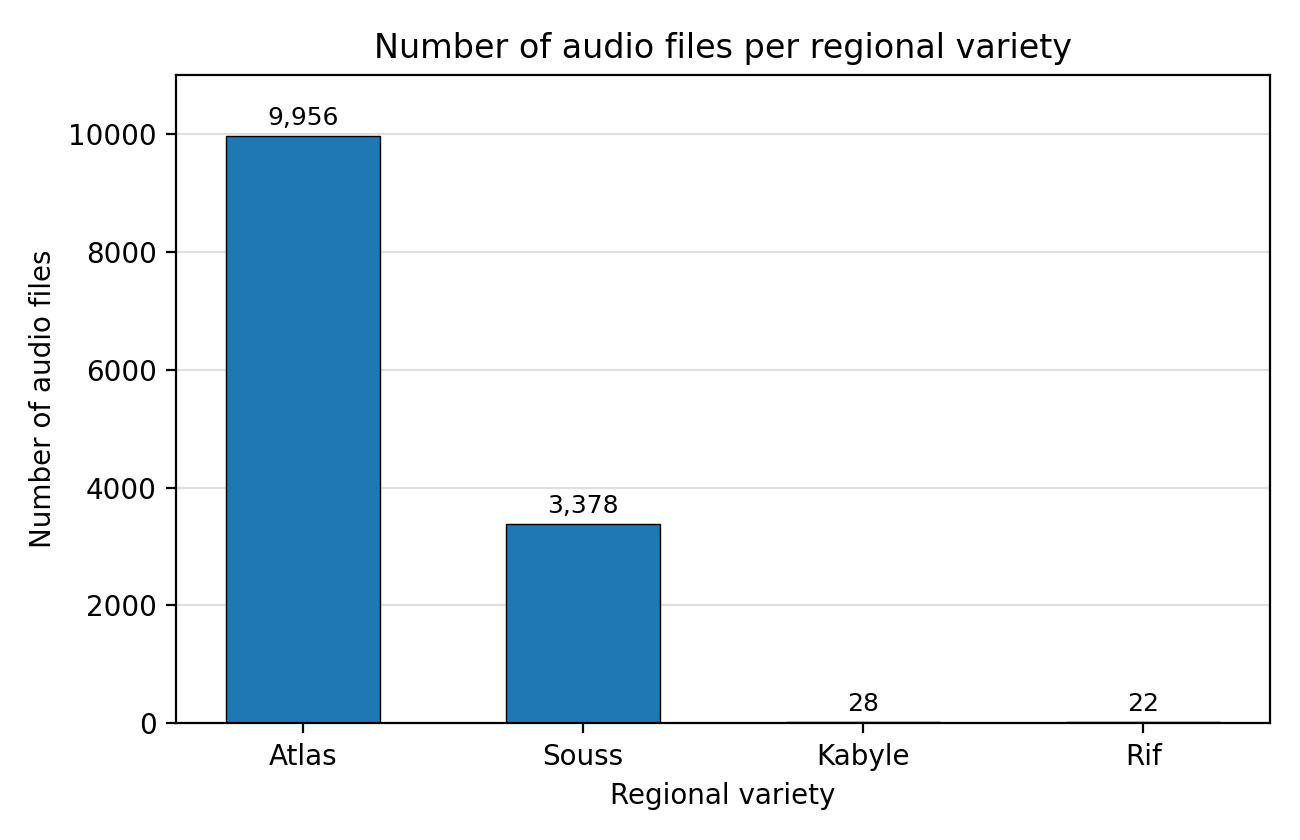}
\caption{Number of audio files per regional variety.}\label{fig:accent_dist}
\end{figure}

\begin{figure}[ht]
\centering
\includegraphics[width=0.72\linewidth]{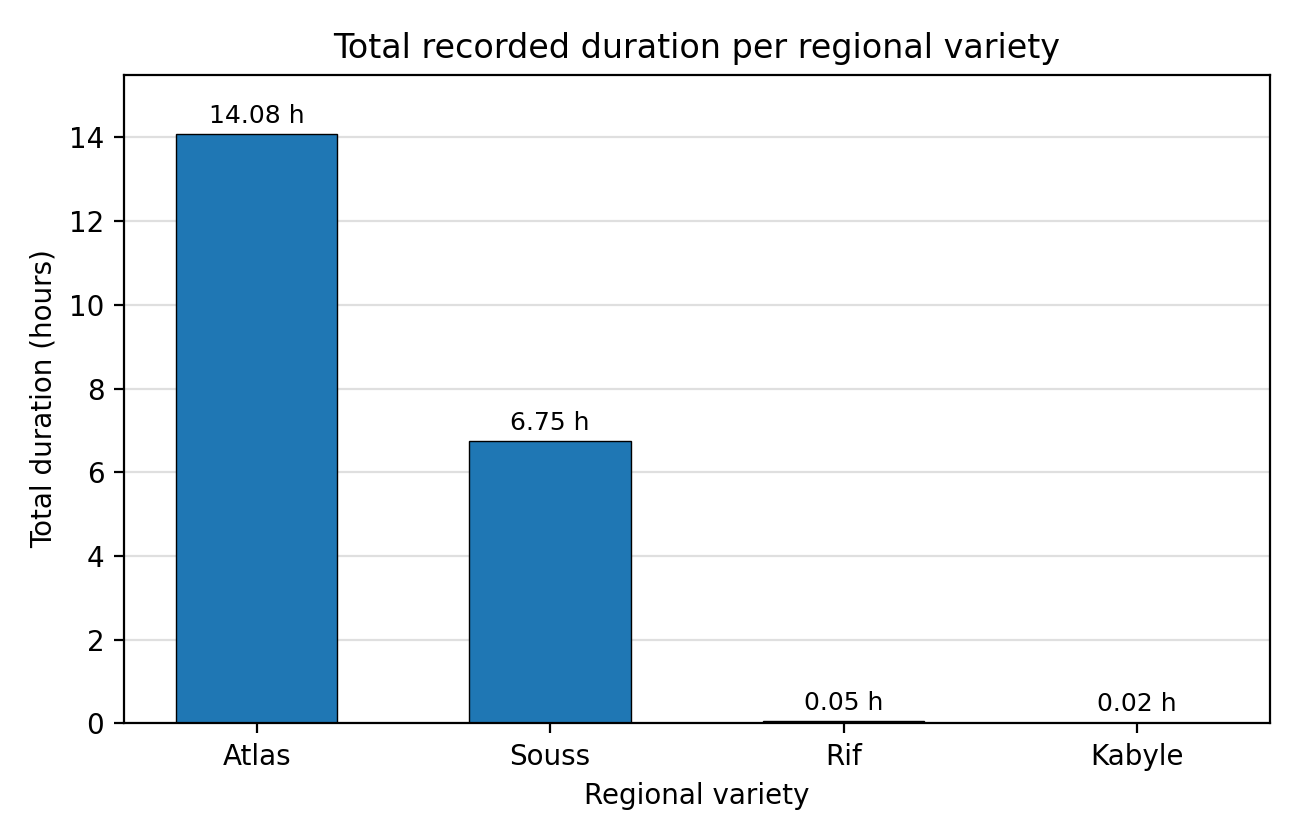}
\caption{Total recorded duration (hours) per regional variety.}\label{fig:accent_duration}
\end{figure}

Tables~\ref{tab:accent_dist} and~\ref{tab:accent_duration}, together with Figs.~\ref{fig:accent_dist} and~\ref{fig:accent_duration}, report the number of files and the cumulative duration for each regional variety. The Atlas variety represents 9{,}956 files (14.08\,h) and the Souss variety 3{,}378 files (6.75\,h). Two small subsets, Kabyle (28 files, 0.02\,h) and Rif (22 files, 0.05\,h), are also included.

\begin{table}[ht]
\caption{Number of audio files per regional variety.}\label{tab:accent_dist}
\begin{tabular}{@{}lrr@{}}
\toprule
Regional variety & Number of files & Percentage \\
\midrule
Atlas & 9{,}956 & 74.39\% \\
Souss & 3{,}378 & 25.24\% \\
Kabyle & 28 & 0.21\% \\
Rif & 22 & 0.16\% \\
\midrule
Total & 13{,}384 & 100\% \\
\botrule
\end{tabular}
\end{table}

\begin{table}[ht]
\caption{Total recorded duration per regional variety.}\label{tab:accent_duration}
\begin{tabular}{@{}lrr@{}}
\toprule
Regional variety & Total duration (hours) & Percentage \\
\midrule
Atlas & 14.08 & 67.61\% \\
Souss & 6.75 & 32.42\% \\
Rif & 0.05 & 0.24\% \\
Kabyle & 0.02 & 0.10\% \\
\midrule
Total & 20.90 & 100\% \\
\botrule
\end{tabular}
\end{table}

\section{Methods}\label{sec:methods}

\subsection{Overview of the collection pipeline}\label{subsec:pipeline}

\begin{figure}[ht]
\centering
\includegraphics[width=0.92\linewidth]{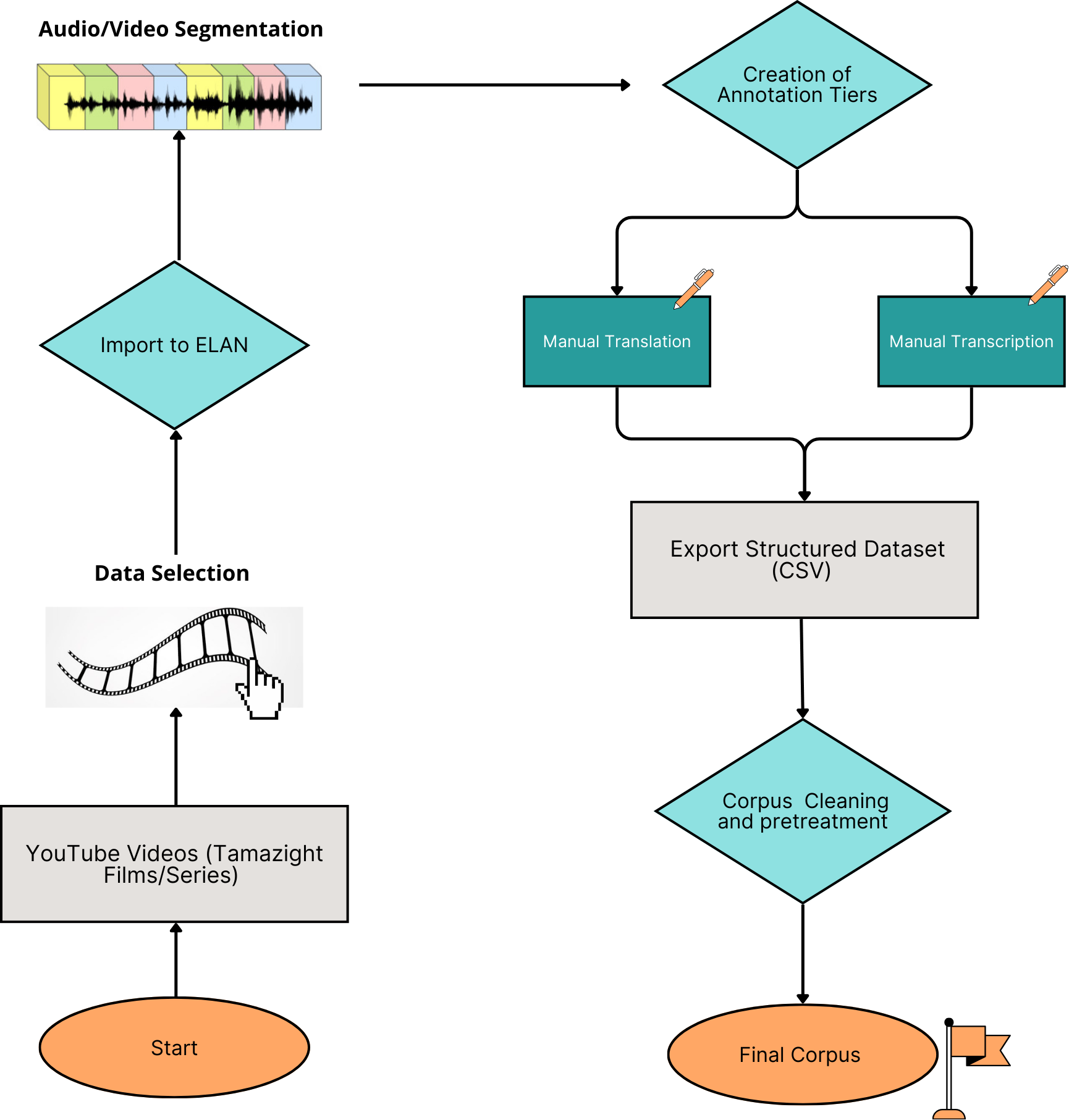}
\caption{Overall workflow of corpus collection, segmentation and annotation.}\label{fig:workflow}
\end{figure}

The corpus was built from two complementary sources, both channelled through the TutlAit web application (Fig.~\ref{fig:workflow}): (i) recordings and transcriptions produced by volunteer native speakers directly on the platform, and (ii) segments extracted from freely accessible Tamazight audiovisual media, annotated offline with ELAN and imported in bulk. All contributions went through the same server-side normalisation, integrity checks and human validation before inclusion.

\subsection{The TutlAit crowdsourcing platform}\label{subsec:platform}

\begin{figure}[ht]
\centering
\includegraphics[width=0.78\linewidth]{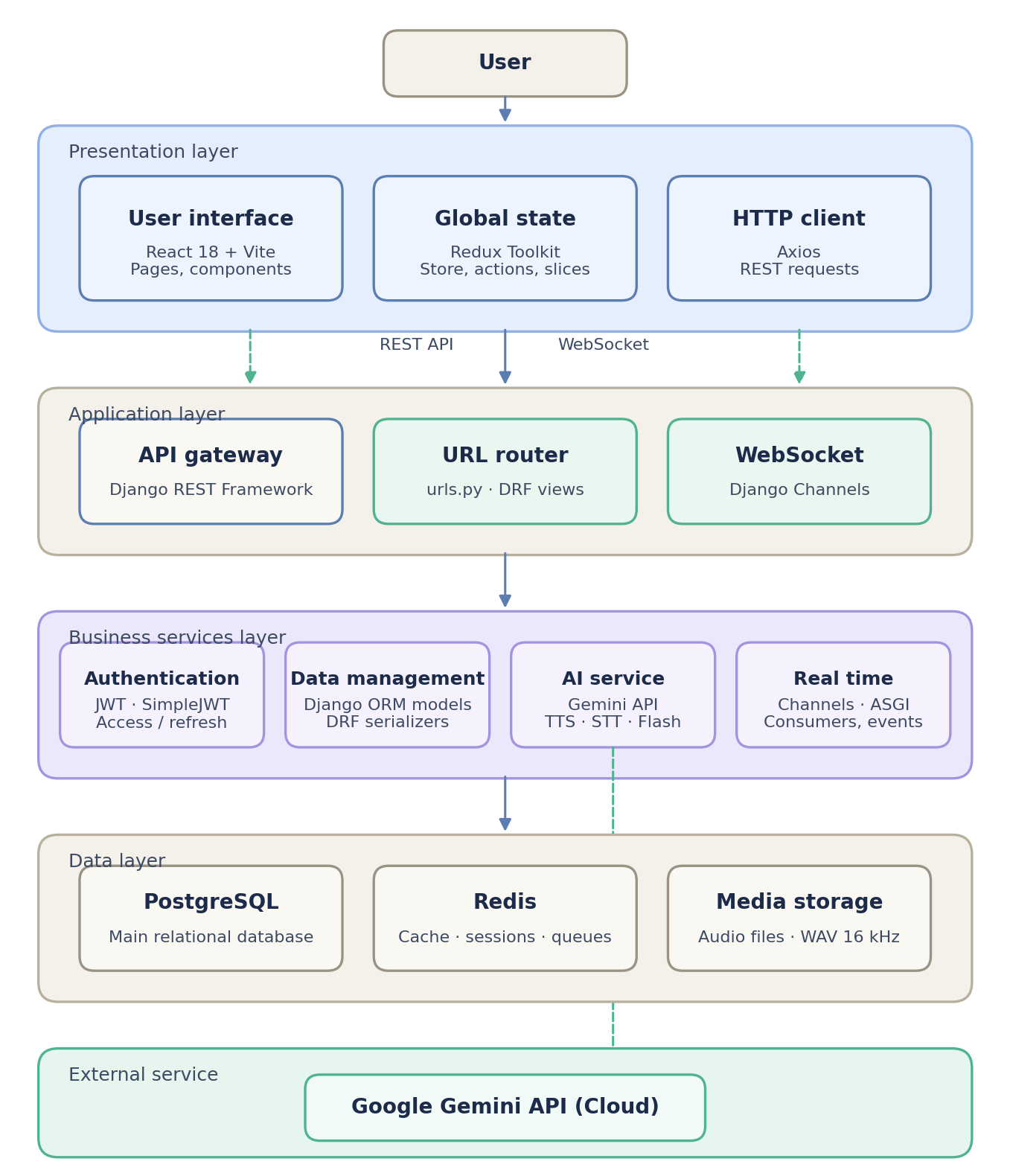}
\caption{General architecture of the TutlAit platform.}\label{fig:archi}
\end{figure}

TutlAit is a full-stack web application designed for building speech--text corpora for Moroccan Tamazight. Its architecture (Fig.~\ref{fig:archi}) is organised in four layers.
\begin{itemize}
\item \textbf{Front-end.} A single-page application built with React 18~\cite{facebook2013react} and bundled with Vite. The interface is available in French, Arabic and English, supports light and dark themes, manages global state with Redux Toolkit and communicates with the back-end through Axios with JSON Web Token (JWT) interceptors~\cite{jones2015jwt}.
\item \textbf{Back-end.} A Django 5 server exposing a RESTful JSON API through Django REST Framework~\cite{christie2011drf}, secured by JWT authentication. It is organised in four Django applications: \texttt{accounts} (users, profiles, roles), \texttt{annotations} (annotation logic, statistics, bulk import), \texttt{translations} (optional AI-assisted translation, text-to-speech and speech-to-text helpers) and \texttt{chat} (administrator--volunteer messaging).
\item \textbf{Persistence.} Metadata and annotations are stored in a PostgreSQL~\cite{postgresql2023} relational database (eleven main tables, including \texttt{CustomUser}, \texttt{AudioSample}, \texttt{TextSample} and \texttt{Annotation}). Audio files are stored on the server file system in a structured media tree.
\item \textbf{Audio processing.} Every uploaded recording is converted server-side to 16\,kHz mono WAV with Pydub and FFmpeg.
\end{itemize}

During the collection campaign the application was deployed behind a Caddy reverse proxy unifying the React front-end and the Django back-end under a single entry point, and exposed on the Internet through an HTTPS tunnel whose URL was shared in the recruitment posts.

\subsection{Participant recruitment and registration}\label{subsec:recruitment}

Volunteers were recruited through targeted campaigns on LinkedIn and Instagram, with communication material adapted to each linguistic community, in order to reach native speakers of the Souss, Rif and Atlas varieties across Morocco. Each volunteer created an account by providing a name, an e-mail address (verified before activation), age, region of origin and institutional affiliation, and selected a regional variety from a controlled vocabulary: Tachelhit (Souss), Tarifit (Rif), Tamazight of the Atlas, and several Algerian varieties (for example Kabyle). These attributes are attached to every contribution, which is how the accent labels were obtained. Only anonymised attributes (hashed speaker identifier, gender, age group, region, variety) are released.

The work involves human participants who voluntarily contributed speech and text. At registration, every participant was informed of the purpose of the collection and of the intended public release of the anonymised recordings and texts, and gave informed consent by accepting the platform's terms of participation. Participants could request deletion of their contributions at any time. Names and e-mail addresses are not included in the release. Media-derived segments were taken from freely accessible public content and contain no personally identifying metadata. Social networks were used only to publish recruitment announcements.

\subsection{Annotation workflows}\label{subsec:workflows}

\begin{figure}[ht]
\centering
\includegraphics[width=0.48\linewidth]{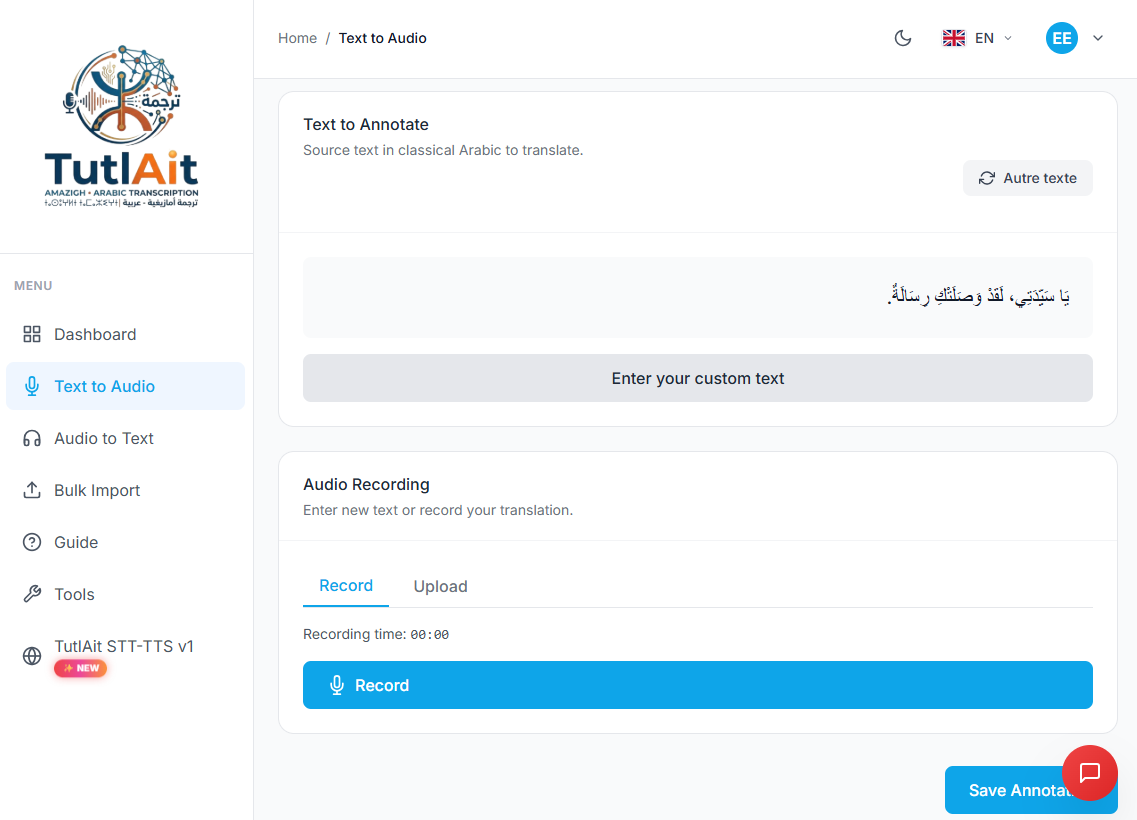}\hfill
\includegraphics[width=0.48\linewidth]{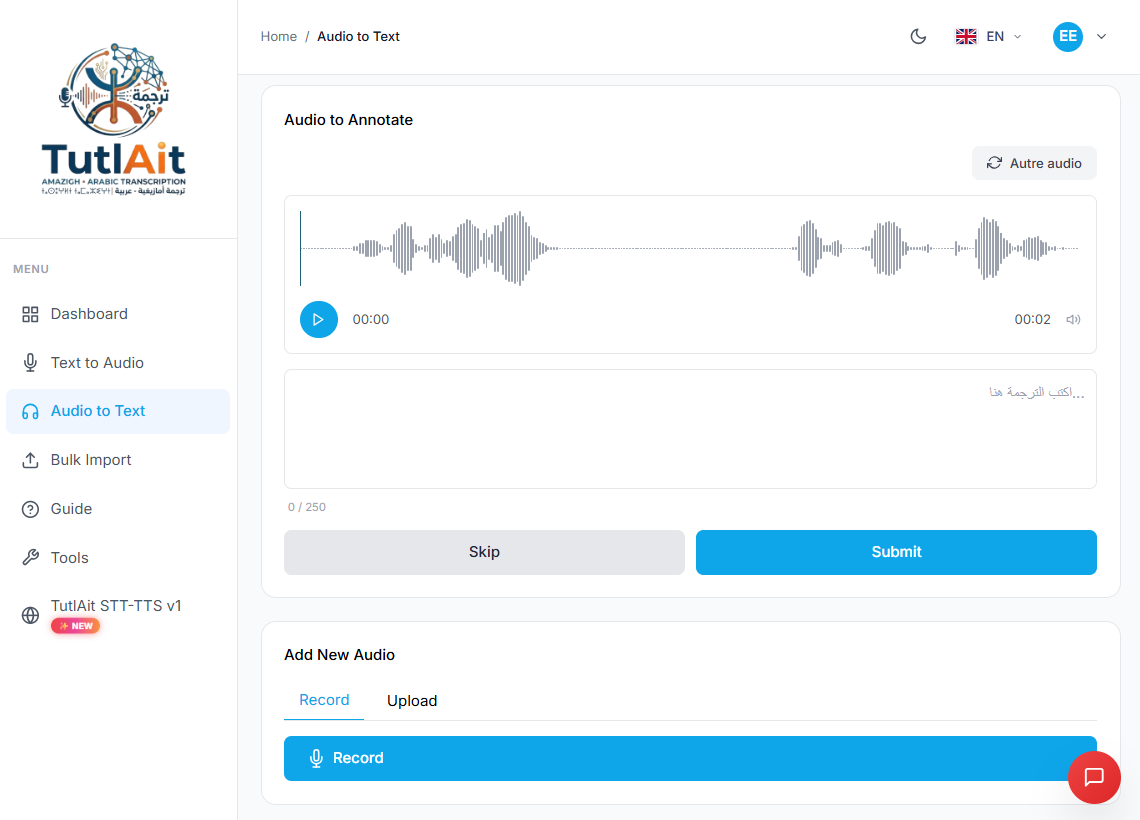}
\caption{TutlAit annotation interfaces. Left: Text-to-Audio. Right: Audio-to-Text.}\label{fig:screens}
\end{figure}

Once logged in, a volunteer is assigned tasks in one of two modes (Fig.~\ref{fig:screens}).
\begin{itemize}
\item \textbf{Text-to-Audio.} A sentence in Modern Standard Arabic is displayed. The volunteer records, in the browser, its oral rendering in his or her Tamazight variety. The displayed Arabic sentence becomes the \texttt{text} field of the pair.
\item \textbf{Audio-to-Text.} A Tamazight excerpt selected according to the volunteer's declared variety is played. The volunteer types its transcription in Arabic, which becomes the \texttt{text} field.
\end{itemize}

An optional assistant based on the Google Gemini API is available within usage quotas to pre-fill a translation or transcription. Every suggestion must be reviewed and confirmed manually before submission. Submitted annotations enter a pending-review state.

\subsection{Media-derived segments and bulk import}\label{subsec:bulk}

A complementary set of Tamazight speech segments was obtained from freely accessible audiovisual productions (films, television series and other media, mainly from YouTube). Recordings were segmented at utterance boundaries and annotated with ELAN 6.8~\cite{elan2024} using three parallel tiers: the time-aligned audio segment, a Latin-script Tamazight transcription, and a manual translation into Modern Standard Arabic. Annotations were exported to CSV and imported through a bulk pipeline that accepts a CSV/XLSX index together with a ZIP archive of audio files. Each row is validated atomically, and the corresponding annotations are created with the \texttt{bulk\_import} mode flag.

\subsection{Quality control and data integrity}\label{subsec:qc}

\begin{figure}[ht]
\centering
\includegraphics[width=0.9\linewidth]{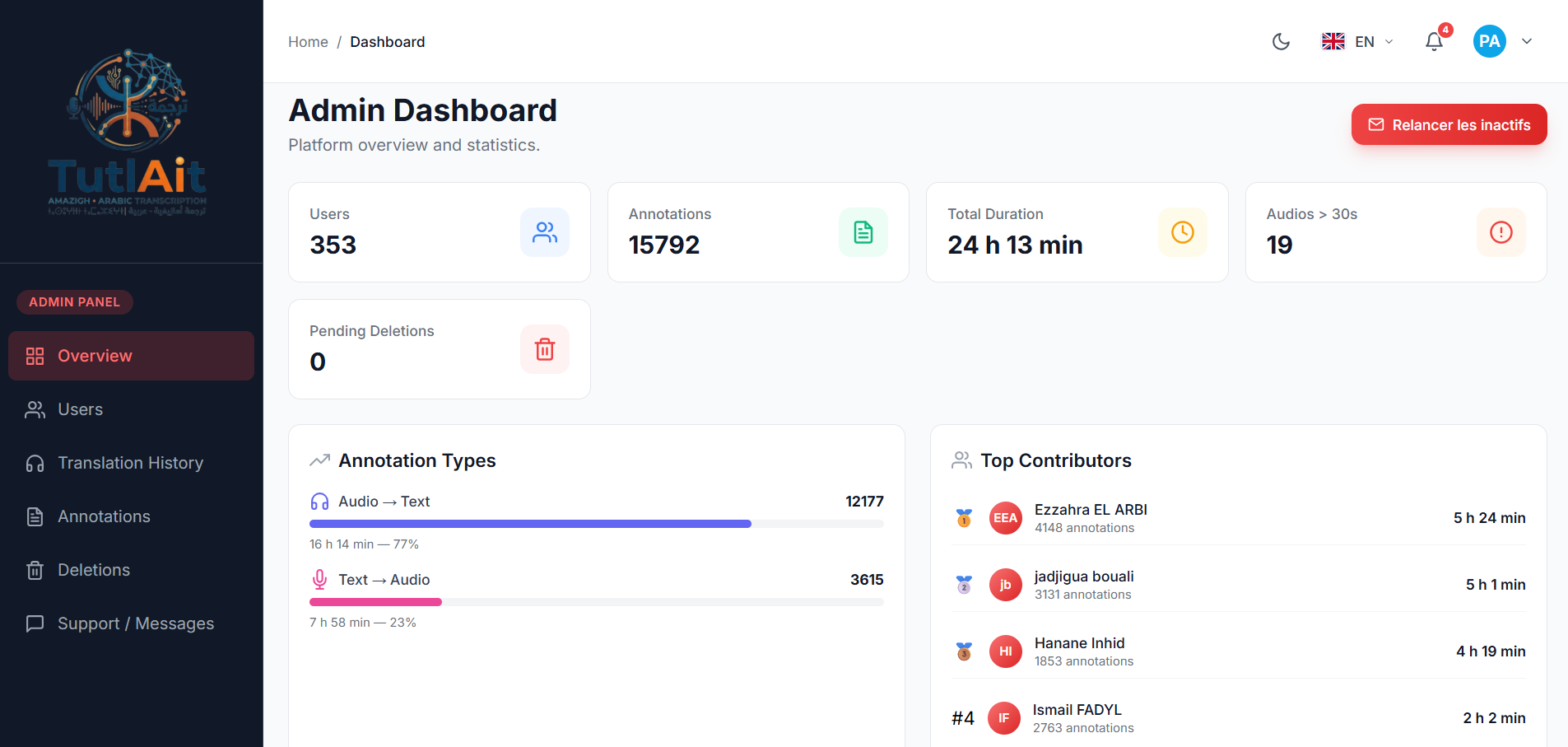}
\caption{TutlAit administrator dashboard used for review and collection statistics.}\label{fig:admin}
\end{figure}

The following mechanisms were applied to every contribution.
\begin{itemize}
\item \textbf{Format normalisation:} conversion to 16\,kHz, mono, 16-bit PCM WAV with FFmpeg/Pydub.
\item \textbf{Duplicate detection:} a SHA-256 hash~\cite{nist2015sha} is computed for each uploaded file; uploads whose hash already exists are rejected.
\item \textbf{Duration bounds:} server-side validation enforces minimum and maximum recording durations at upload time.
\item \textbf{File integrity:} each file is opened and decoded to verify that it is readable and not corrupted.
\item \textbf{Human review:} an administrator dashboard (Fig.~\ref{fig:admin}) lists pending contributions; each item is validated or rejected. Only validated items are released.
\item \textbf{Soft deletion:} deletions requested by participants use an \texttt{is\_exist} flag so that the audit trail is preserved. Soft-deleted items are excluded from the release.
\end{itemize}

\subsection{Export of the released dataset}\label{subsec:export}

The released files were produced by a Python export script run against the TutlAit PostgreSQL database. The script selects all annotations with validated status and \texttt{is\_exist} true, copies the corresponding WAV files into the variety-specific sub-folders, recomputes duration and SHA-256, replaces account identifiers by salted hashes, and writes \texttt{metadata.csv} and \texttt{metadata.xlsx}. The descriptive statistics and figures were computed from this index with pandas and Matplotlib.

\section{Limitations}\label{sec:limits}

The distribution across regional varieties is strongly unbalanced: the Atlas variety accounts for about two thirds of the recorded duration and Souss for about one third, whereas the Rif (22 files, 0.05\,h) and Kabyle (28 files, 0.02\,h) subsets are too small to be used on their own. Accent labels reflect the variety self-declared by each speaker at registration and were not independently verified by a linguist. Recordings were made by volunteers on their own devices in uncontrolled environments, so microphone quality, background noise and loudness vary between speakers. The text field is Modern Standard Arabic (a translation of the spoken Tamazight) rather than a Tamazight transcription in Tifinagh or Latin script, which restricts some phonetic uses of the data. Nineteen files exceed 30 seconds and one file is shorter than 0.2 seconds; users targeting fixed-window models may wish to filter on \texttt{duration\_s}. Speaker metadata (age group, gender, region) are self-reported and were not checked.

\backmatter

\bmhead{Acknowledgements}

\begin{figure}[H]
\centering
\includegraphics[width=0.55\linewidth]{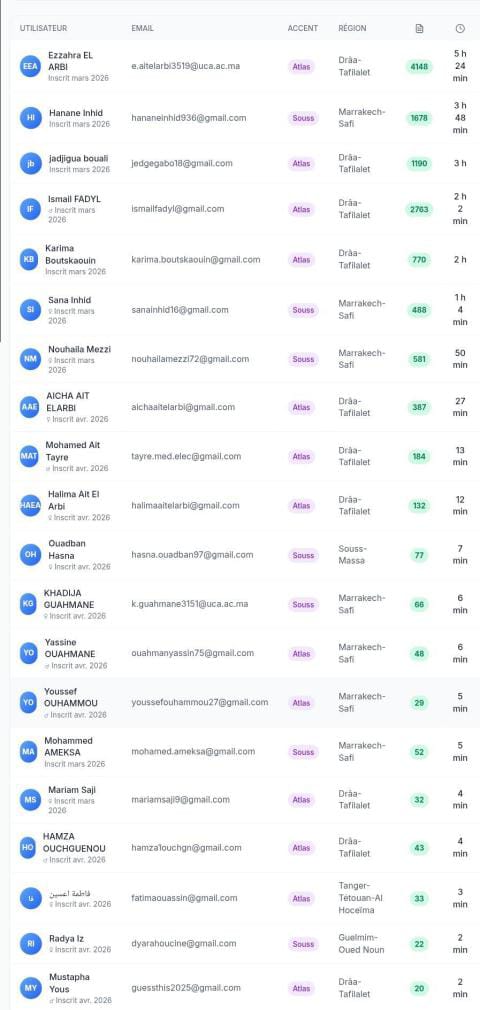}
\caption{Contributors of the TutlAit v1 collection.}\label{fig:contributors}
\end{figure}

The authors thank all the volunteer native speakers who contributed recordings and transcriptions through the TutlAit platform. Figure~\ref{fig:contributors} lists the registered contributors of this first release (TutlAit v1), with the accent and region they declared, the number of annotations and the recorded duration.

\section*{Declarations}

\begin{itemize}
\item \textbf{Conflict of interest.} The authors declare that they have no known competing financial interests or personal relationships that could have appeared to influence the work reported in this paper.
\item \textbf{Ethics approval and consent to participate.} Participants gave informed consent through the platform's terms of participation before contributing. They were informed that anonymised recordings and texts would be released publicly and could request deletion of their contributions at any time. The work does not involve animal experiments.
\item \textbf{Consent for publication.} Not applicable. No identifying personal data are published.
\item \textbf{Data availability.} Direct link: \url{https://huggingface.co/datasets/Ma-OpenHub/Tutlait-v1}. New validated recordings are added to this same dataset.
\end{itemize}

\bibliography{references}

\end{document}